%% file: main.tex
\documentclass{article} 
\usepackage{iclr2026_conference,times}

\input{math_commands}

\usepackage{hyperref}
\usepackage{url}
\usepackage{graphicx}
\usepackage{subcaption}
\usepackage{multirow}

\usepackage{booktabs} 
\usepackage{geometry}
\usepackage{amsmath}  

\title{Non-Stationarity in the Embedding Space\\of Time Series Foundation Models}

\author{Jinmyeong Choi, Brad Shook, Artur Dubrawski \\
Auton Lab\\
Robotics Institute\\
Carnegie Mellon University\\
Pittsburgh, PA 15213, USA \\
\texttt{\{jinmyeoc, bshook, awd\}@andrew.cmu.edu} \\
}

\iclrfinalcopy
\track{Research}

\begin{document}

\maketitle

\begin{abstract}
Time series foundation models (TSFMs) are widely used as generic feature extractors, yet the notion of non‑stationarity in their embedding spaces remains poorly understood. Recent work often conflates non‑stationarity with distribution shift, blurring distinctions fundamental to classical time‑series analysis and long‑standing methodologies such as statistical process control (SPC). In SPC, non‑stationarity signals a process leaving a stable regime—via shifts in mean, variance, or emerging trends—and detecting such departures is central to quality monitoring and change‑point analysis. Motivated by this diagnostic tradition, we study how different forms of distributional non‑stationarity—mean shifts, variance changes, and linear trends—become linearly accessible in TSFM embedding spaces under controlled conditions. We further examine temporal non‑stationarity arising from persistence, which reflects violations of weak stationarity due to long‑memory or near‑unit‑root behavior rather than explicit distributional shifts. By sweeping shift strength and probing multiple TSFMs, we find that embedding‑space detectability of non‑stationarity degrades smoothly and that different models exhibit distinct, model‑specific failure modes.

\end{abstract}
\vspace{-5mm}

\input{main/1_introduction}

\input{main/2_nonstationarity}

\input{main/4_discussion}

\subsubsection*{Acknowledgments}
This work was supported by Institute of Information \& communications Technology Planning \& Evaluation (IITP) grant funded by the Korea government(MSIT) (RS-2022-00143911, AI Excellence Global Innovative Leader Education Program). This work has been partially supported by the National Science Foundation (awards 2406231 and 2427948).

\bibliography{iclr2026_conference}
\bibliographystyle{iclr2026_conference}

\newpage
\appendix
\input{appendix/1_data_generating}

\input{appendix/2_model_explanation}

\input{appendix/4_results}
\input{appendix/3_background}

\end{document}

%% file: math_commands.tex
\usepackage{amsmath,amsfonts,bm}

\def\eqref#1{equation~\ref{#1}}

\def\1{\bm{1}}

\DeclareMathAlphabet{\mathsfit}{\encodingdefault}{\sfdefault}{m}{sl}
\SetMathAlphabet{\mathsfit}{bold}{\encodingdefault}{\sfdefault}{bx}{n}



%% file: main/1_introduction.tex
\section{Introduction}
\label{sec:introduction}

Time series foundation models (TSFMs) are increasingly used as generic feature extractors in various downstream tasks \citep{ansari2024chronos, ansari2025chronos2, goswami2024moment, TiRex, talukder2024totem, survey_tsfm}.
Despite their growing adoption, the interpretation of \emph{non-‐stationarity} within their embedding spaces remains unclear. 
In much of the recent literature, non-‐stationarity is treated synonymously with \emph{distribution shift}, typically referring to changes in mean, variance, or the presence of trends. However, classical time‑series theory distinguishes these changes from violations of weak stationarity arising from persistent temporal dependence~\citep{kim2021reversible, liu2022non, dishts}.

This distinction is grounded historically in Statistical Process Control (SPC), where the goal is to detect departures from an in-control, stable process. In SPC, \emph{mean shifts}, \emph{variance inflation}, and \emph{trends} are treated as distinct structural changes with different operational implications, and canonical SPC methods were explicitly designed to trade off sensitivity profiles and average run lengths~\citep{MontgomerySQC7}.

In classical statistics, non-stationarity is defined through violations of weak stationarity, such as the presence of unit roots or time-varying autocovariance.
In contrast, many modern approaches treat non-stationarity as a nuisance factor associated with changing marginal statistics, motivating normalization and stabilization techniques.
These differing perspectives raise a fundamental question: \emph{what aspects of non-stationarity are actually preserved or suppressed in the embedding spaces of TSFMs?}

We adopt a diagnostic approach to find the answer: under controlled autoregressive conditions, we probe how \emph{distributional} non‑stationarity—mean and variance shifts, linear trends—becomes accessible in TSFM embeddings as shift magnitude varies, and we contrast these with non‑stationarity induced by \emph{temporal persistence}. 
By systematically reducing the extent of shift and comparing multiple foundation models, we show that the appearance of non-stationarity in embedding spaces diminishes smoothly rather than abruptly.

%% file: main/2_nonstationarity.tex
\input{figures/fig_distribution}

\section{Probing Non-Stationarity in TSFM Embeddings}


We distinguish two notions of non-stationarity that are often conflated in embedding-based analyses: distributional non-stationarity and temporal non-stationarity.
The first refers to explicit changes in marginal statistics over time, such as shifts in mean, changes in variance, or the presence of deterministic trends.
These effects are commonly treated as distribution shift.
In embedding space, a central diagnostic question is whether such deviations remain linearly accessible from learned representations.

On the other hand, temporal non-stationarity is defined in classical time series analysis as a violation of weak stationarity.
For the AR(1) process
\[
x_t = \mu + \phi (x_{t-1} - \mu) + \varepsilon_t,
\quad \varepsilon_t \sim \mathcal{N}(0, \sigma^2),
\]
weak stationarity holds for $|\phi| < 1$ and breaks down at the unit-root boundary $\phi = 1$ \citep{Cryer_Chan_2008}.
Importantly, increasing persistence does not imply large distributional shifts within finite windows.

We use controlled AR(1) settings to analyze how these two forms of non-stationarity manifest in TSFM embeddings.
We first examine distributional deviations and then contrast them with persistence-based effects.

\paragraph{Distributional Non-Stationarity.}
Figure~\ref{fig:distributional} visualizes representative AR(1) windows under stationary, mean shift, variance shift, and trend conditions, together with their Chronos2 embeddings.
In the raw data space, the shift types are visually distinct, and they are detectable by standard SPC techniques.
In the embedding space, the data observed before and after the shift form distinct clusters, indicating that distributional deviations are preserved.
However, visualization alone does not determine whether such information is linearly accessible.
To quantify this, we conduct linear probing under progressively weaker shift magnitudes.


We generate window-level AR(1) sequences ($L=128$) under four shift types: stationary (no shift), mean shift, variance shift, and trend.
To control task difficulty, we introduce a shift-strength parameter $s \in (0,1]$ that scales the magnitude of effect.
%
%
For each data window, we obtain embeddings from Chronos2, MOMENT, TOTEM and train a multinomial linear classifier to predict the type of shift~\citep{ansari2025chronos2, goswami2024moment, talukder2024totem}. 
For comparison, we include two logistic regression baselines: \textit{Stats-LR} uses only summary statistics as input features, while \textit{Stats+Dynamics-LR} includes additional dynamical characteristics.
%
The results are reported using Macro-F1 averaged over 5 random seeds. Detailed data generation parameters are described in Appendix~\ref{app:data_gen}, further modeling details are in Appendix~\ref{sec:appendix_models}, and full numerical results are provided in Appendix~\ref{app:results}.
\input{figures/figure4}


As expected, overall decodability diminishes as the extent of shift is reduced.
Figure~\ref{fig:macro_f1_two_panel} shows Macro-F1 as a function of shift strength.
When shifts are strong ($s=1.0$), all models—including statistics-based baselines—achieve high separability across shift types.
In particular, \textit{Stats-LR} attains near-perfect performance, indicating that large distributional shifts are easily detectable using marginal statistics alone.

As the shift magnitude decreases, its detection performance degrades gradually at 
rates that vary across models.
While statistics-based baselines degrade rapidly, Chronos2 retains the highest separability in weaker regimes, followed by MOMENT, whereas TOTEM exhibits a markedly earlier collapse. Table~\ref{tab:macro_f1_integrated} in the appendix quantifies these gaps in detail.
%
These results suggest that pretrained TSFM representations capture weak distributional deviations that are not equally reflected in the handcrafted statistical features.

\input{figures/fig_temporal}

\paragraph{Temporal Non-Stationarity}
%
Figure~\ref{fig:temporal_a} visualizes AR(1) windows with increasing $\phi$ and their corresponding Chronos2 embeddings. As $\phi$ increases, the raw signals exhibit stronger temporal dependence and longer memory. In contrast, the embedding representation evolves smoothly, without a sharp transition at the unit-root boundary ($\phi = 1$).

To quantify this behavior, we continuously vary $\phi$ and measure embedding discrepancies using cosine distance (see Figure~\ref{fig:temporal_b} and \ref{fig:temporal_c}).
Across all models, embedding distances increase monotonically with $\phi$, indicating that persistence is encoded as a graded factor. This interpretation is further supported by the linear probing results (Appendix~\ref{app:phi_regression}), which show that $\phi$ can be predicted from embeddings with high accuracy, confirming that persistence information is explicitly encoded rather than merely reflected in distance metrics.
In particular, no discontinuity is observed at the unit-root boundary, although classical definitions treat $\phi=1$ as a qualitative change in stationarity. These results suggest that TSFM embeddings do not reflect the classical stationarity boundary in a discrete manner.
Instead, persistence is represented as a continuous dimension of the temporal structure. Model-specific failure modes are further analyzed via confusion matrices (Appendix \ref{app:confusion}).





%% file: figures/fig_distribution.tex
\begin{figure}[t]
\centering
\begin{subfigure}[c]{0.35\linewidth}
    \centering
    \includegraphics[width=\linewidth]{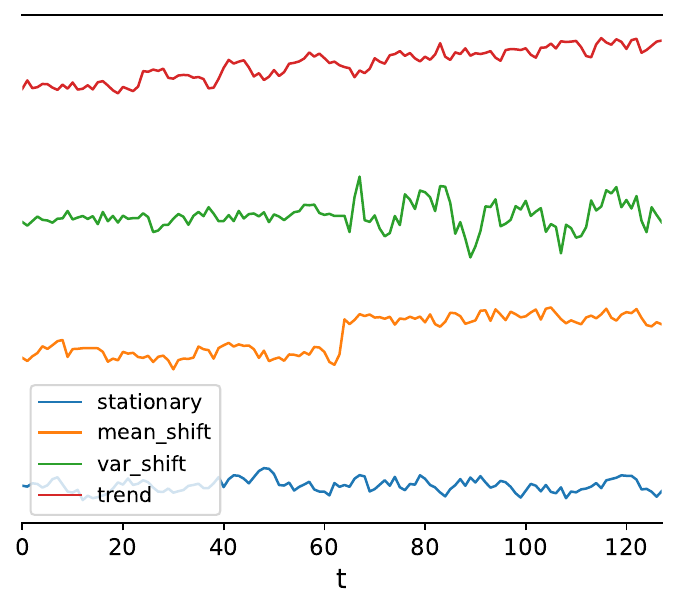}
    \caption{Representative AR(1) windows under stationary, mean shift, variance shift, and trend.}
    \label{fig:distributional_a}
\end{subfigure}
\hspace{0.5cm}
\begin{subfigure}[c]{0.4\linewidth}
    \centering
    \includegraphics[width=\linewidth]{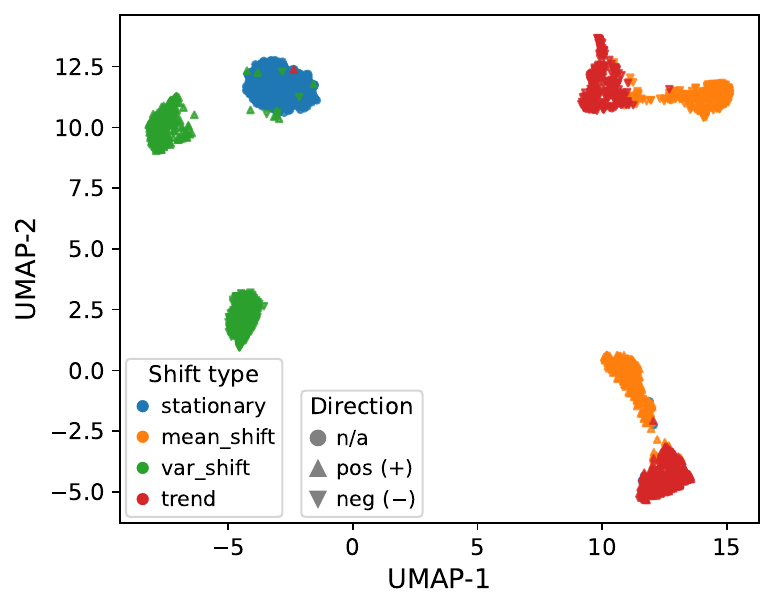}
    \caption{Two-dimensional UMAP projection of Chronos2 embeddings for the same windows.}
    \label{fig:distributional_b}
\end{subfigure}

\caption{
Distributional non-stationarity in raw and embedding space (Chronos2).
}
\label{fig:distributional}
\end{figure}
\vspace{-0.8em}

%% file: figures/figure4.tex

\begin{figure}[t]
\centering
\begin{subfigure}{0.48\linewidth}
    \centering
    \includegraphics[width=\linewidth]{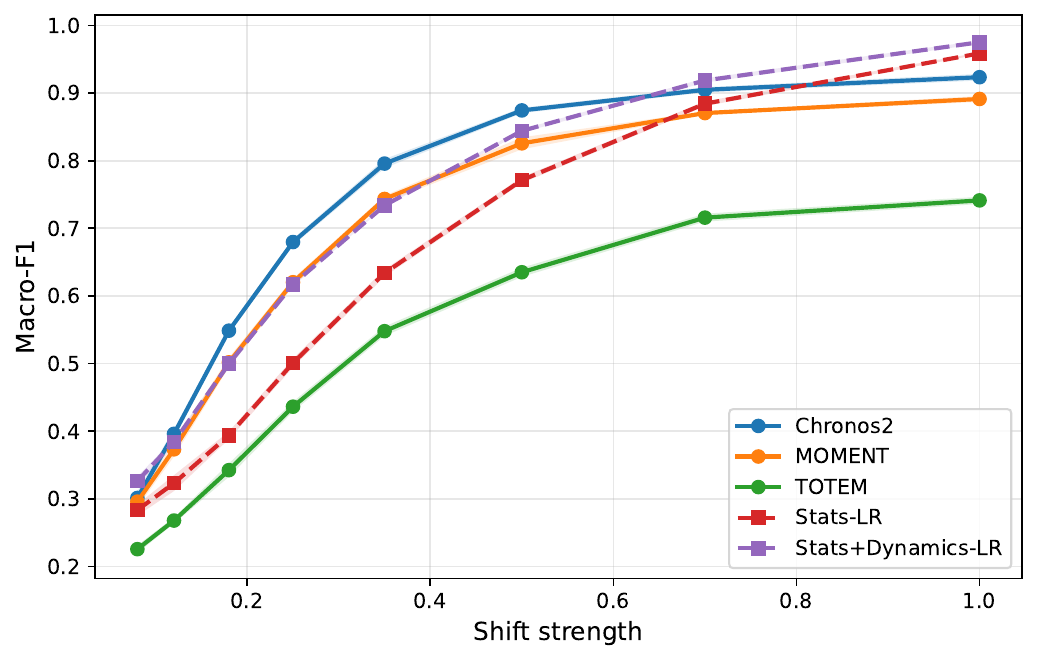}
    \caption{Fixed persistence ($\phi=0.6$).}
    \label{fig:macro_fixed}
\end{subfigure}
\hfill
\begin{subfigure}{0.48\linewidth}
    \centering
    \includegraphics[width=\linewidth]{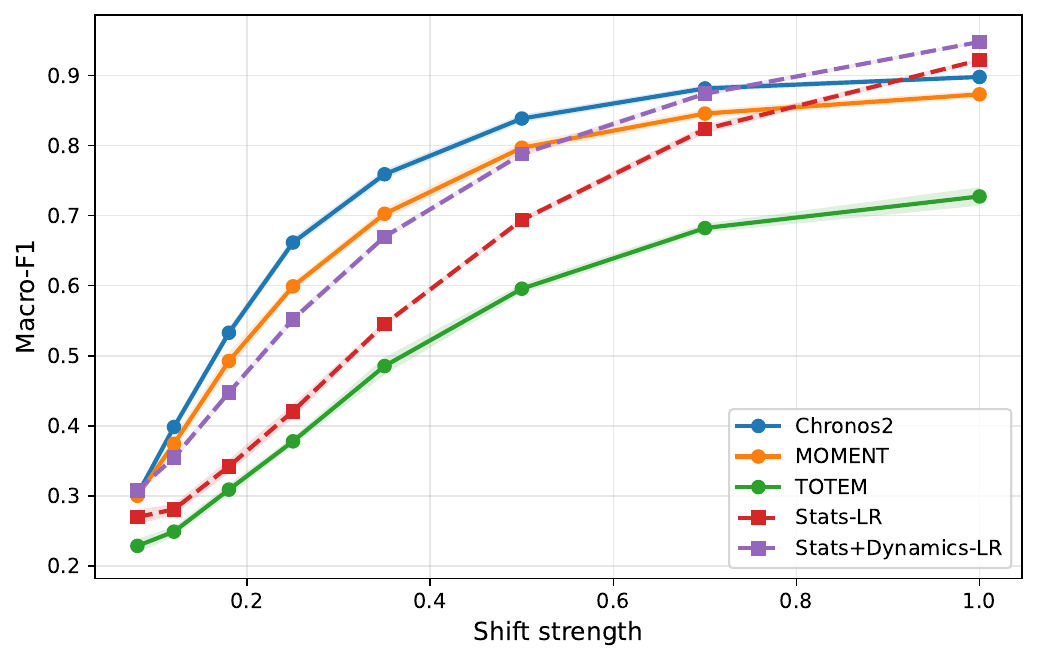}
    \caption{Random persistence ($\phi \sim U(0.3,0.9)$).}
    \label{fig:macro_random}
\end{subfigure}

\caption{
Macro-F1 as a function of shift strength.
Strong shifts are easily detected by both TSFM embeddings and statistics-based baselines.
As shift magnitude decreases, separability degrades smoothly, with statistics-based baselines degrading more rapidly than Chronos2 and MOMENT.
The qualitative behavior remains unchanged under random persistence, indicating that shift-type decodability is driven primarily by distributional structure rather than autoregressive persistence.
}
\label{fig:macro_f1_two_panel}
\end{figure}

%% file: figures/fig_temporal.tex

\begin{figure}[t]
    \centering
    \begin{subfigure}[b]{0.32\linewidth}
        \centering
        \includegraphics[width=\linewidth]{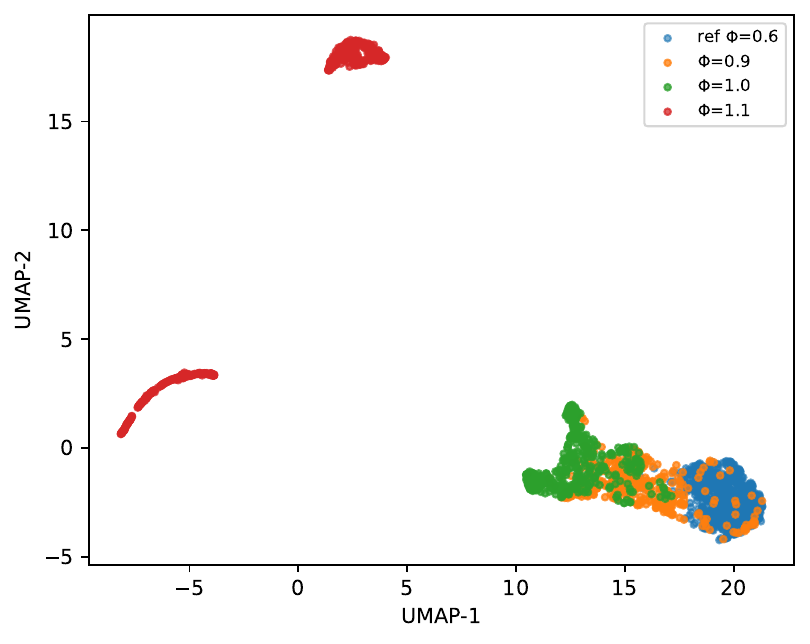}
        \caption{Chronos2 Embeddings (UMAP)}
        \label{fig:temporal_a}
    \end{subfigure}
    \hfill 
    \begin{subfigure}[b]{0.32\linewidth}
        \centering
        \includegraphics[width=\linewidth]{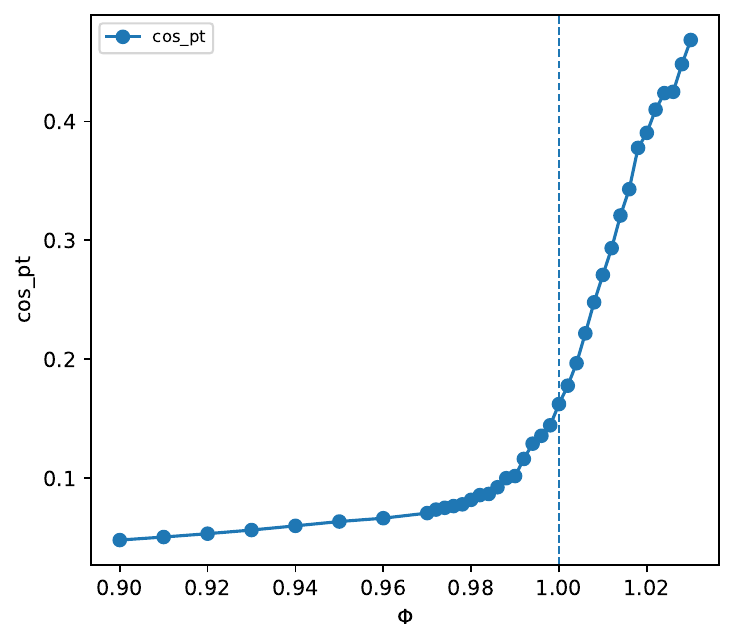}
        \caption{$\phi$-sweep (cosine, raw)}
        \label{fig:temporal_b}
    \end{subfigure}
    \hfill
    \begin{subfigure}[b]{0.32\linewidth}
        \centering
        \includegraphics[width=\linewidth]{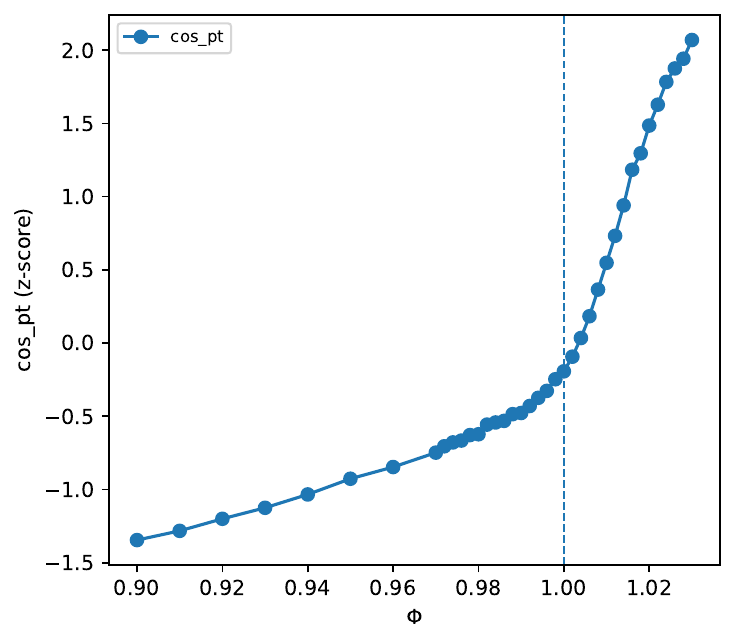}
        \caption{$\phi$-sweep (cosine, z-score)}
        \label{fig:temporal_c}
    \end{subfigure}
    
    \caption{
        Temporal non-stationarity in TSFM embeddings under an AR(1) persistence sweep.
        (a) Chronos2 embeddings shift smoothly with increasing $\phi$, without a sharp transition at $\phi=1$.
        (b) Cosine discrepancy (raw) increases monotonically with persistence.
        (c) The same trend persists after z-score normalization, indicating that the TSFM embeddings capture temporal dependence beyond scale effects.
    }
    \label{fig:temporal}
\end{figure}

%% file: main/4_discussion.tex
\section{Discussion and Conclusion}
\label{sec:discussion}

Our results suggest that the appearance of non-stationarity in TSFM embedding spaces is neither abrupt nor model-agnostic.
Instead, its linear accessibility depends on the type and magnitude of the underlying shift, as well as on the inductive biases of the representation. Detectability degrades smoothly with reduced magnitude of shift rather than collapsing at a particular boundary.
%
Importantly, different shift types fail in distinct ways.
Chronos2 and MOMENT retain access to variance-based deviations even under weak perturbations, while TOTEM exhibits an early collapse of mean shifts into the stationary class.
These structured failure modes indicate that embedding spaces do not uniformly preserve all forms of relevant information.
Rather, they selectively suppress or retain specific statistical attributes, such as absolute level or scale.

The persistence-based experiments further suggest that non-stationarity cannot be fully characterized in the embedding spaces by unit-root notions alone.
Even when autoregressive coefficients are randomized, qualitative failure patterns persist.
This highlights a gap between classical definitions of non-stationarity and how modern foundation models encode temporal structure that should be investigated further and perhaps pragmatically exploited.

The key takeaway is that TSFMs offer a unified representational interface in which diverse forms of non‑stationarity—mean shifts, variance changes, trends, and persistence‑based deviations—become jointly linearly accessible, enabling lightweight linear probes to recover sensitivity to a broad range of departures from stationarity without relying on bespoke, shift‑specific detectors. 
This reframes non‑stationarity detection as a problem of learned representation rather than a collection of manually engineered statistical tests. While this universality does not replace the rigorous guarantees of SPC, it complements them: TSFM embeddings can front‑end classical detectors by surfacing heterogeneous change types through a shared representation, reducing the operational overhead of maintaining multiple specialized charts.
Overall, this positions TSFMs as promising, domain‑agnostic detectors of structured non‑stationarity, capable of consolidating multiple monitoring tasks into a single embedding space and enabling more scalable approaches to forecasting and change detection in complex real‑world systems.



%% file: appendix/1_data_generating.tex
\section{Data Generating Process}
\label{app:data_gen}

We generate synthetic time series using a controlled AR(1) process to study how distributional and temporal non-stationarity manifest in embedding space.
All experiments are conducted at the window level with sequence length $L=128$. 
Visualization of baseline and shifts are in Figure \ref{fig:ar1_synth_vis}.
\subsection{Baseline AR(1) Process}

The stationary baseline is defined as an AR(1) process
\[
x_t = \mu + \phi (x_{t-1} - \mu) + \varepsilon_t,
\quad
\varepsilon_t \sim \mathcal{N}(0, \sigma^2),
\]
where $\mu=0.5$, $\sigma=0.06$, and $|\phi|<1$ ensures weak stationarity.
Unless otherwise specified, we use $\phi=0.6$.

This baseline defines the \textbf{stationary} class.

\subsection{Distributional Shift Types}

We consider three forms of distributional non-stationarity applied at the window level.

\paragraph{Mean shift.}
The window is split into two halves.
The first half follows the baseline process, while the second half uses a shifted mean:
\[
\mu' = \mu + \Delta_\mu.
\]
This produces a piecewise-constant mean within the window.

\paragraph{Variance shift.}
The window is split into two halves with different innovation variances:
\[
\varepsilon_t \sim
\begin{cases}
\mathcal{N}(0, \sigma_1^2), & t \le L/2, \\
\mathcal{N}(0, \sigma_2^2), & t > L/2.
\end{cases}
\]
The second half continues from the last value of the first half to preserve temporal continuity.

\paragraph{Trend.}
A deterministic linear trend is added to the baseline process:
\[
x_t^{\text{trend}} = x_t + \text{linspace}(0, \alpha, L),
\]
where $\alpha$ controls the slope of the trend.

\subsection{Shift Strength Scaling}

To control task difficulty, we introduce a shift-strength parameter $s \in (0,1]$ that scales the magnitude of all distributional shifts.

\paragraph{Mean shift scaling.}
\[
\Delta_\mu \sim \text{Uniform}(0.2s,\, 0.6s) \cdot \text{sign}.
\]

\paragraph{Trend scaling.}
\[
\alpha \sim \text{Uniform}(0.3s,\, 0.6s) \cdot \text{sign}.
\]

\paragraph{Variance shift scaling.}
Let $\sigma_0 = 0.06$ denote the baseline standard deviation.
We sample
\[
\sigma_{\text{low}} \sim \text{Uniform}(0.03, 0.06), \quad
\sigma_{\text{high}} \sim \text{Uniform}(0.12, 0.20),
\]
and interpolate toward $\sigma_0$ using strength $s$:
\[
\sigma_1 = \sigma_0 + s(\sigma_{\text{low}} - \sigma_0), \quad
\sigma_2 = \sigma_0 + s(\sigma_{\text{high}} - \sigma_0).
\]
As $s \to 0$, both variances converge to $\sigma_0$, making the shift indistinguishable from the stationary class.

\subsection{Temporal Persistence as a Nuisance Factor}

To assess whether persistence drives shift separability, we also consider a nuisance setting where
\[
\phi \sim \text{Uniform}(0.3, 0.9)
\]
is sampled independently for each window while keeping the same distribution across all classes.
This prevents label leakage while testing whether shift-type decodability depends on autoregressive persistence.

\input{figures/phi_raw_samples}

\subsection{Illustration of Temporal Persistence}
\label{app:phi_examples}

To illustrate temporal non-stationarity in the raw signal space, Figure~\ref{fig:phi_raw_examples}
shows representative AR(1) windows generated with increasing values of $\phi$.
When $\phi=0.6$, the process remains in the weakly stationary regime and fluctuates around the mean.
As $\phi$ increases to $0.9$, temporal dependence becomes stronger and the trajectory evolves more slowly.
At the unit-root boundary $\phi=1.0$, the process no longer reverts to the mean and instead exhibits random-walk-like behavior.
For $\phi=1.1$, the process becomes explosive, producing rapidly diverging trajectories.

These examples highlight that temporal non-stationarity differs qualitatively from the distributional shifts considered above.
Rather than introducing an explicit change in mean, variance, or trend within a window, varying $\phi$ changes the persistence structure of the process itself.

\begin{table}[t]
\centering
\caption{
Parameter ranges used for synthetic data generation.
Shift magnitudes are scaled by strength $s \in (0,1]$ to control task difficulty.
}
\label{tab:data_gen_params}
\resizebox{0.95\columnwidth}{!}{
\begin{tabular}{l l l}
\toprule
Category & Parameter & Value / Range \\
\midrule

\multicolumn{3}{l}{\textbf{Baseline AR(1)}} \\
\midrule
 & Window length $L$ & 128 \\
 & Mean $\mu$ & 0.5 \\
 & Innovation std.\ $\sigma$ & 0.06 \\
 & Persistence $\phi$ (fixed) & 0.6 \\
 & Persistence $\phi$ (nuisance) & $\text{Uniform}(0.3, 0.9)$ \\
\midrule

\multicolumn{3}{l}{\textbf{Mean Shift}} \\
\midrule
 & Mean change $\Delta_\mu$ & $\text{Uniform}(0.2s, 0.6s) \cdot \text{sign}$ \\
 & Shift structure & Half-window change \\
\midrule

\multicolumn{3}{l}{\textbf{Variance Shift}} \\
\midrule
 & Baseline std.\ $\sigma_0$ & 0.06 \\
 & Low variance & $\text{Uniform}(0.03, 0.06)$ \\
 & High variance & $\text{Uniform}(0.12, 0.20)$ \\
 & Strength interpolation & $\sigma = \sigma_0 + s(\sigma_{\text{raw}} - \sigma_0)$ \\
 & Shift structure & Half-window change (continuous) \\
\midrule

\multicolumn{3}{l}{\textbf{Trend}} \\
\midrule
 & Slope $\alpha$ & $\text{Uniform}(0.3s, 0.6s) \cdot \text{sign}$ \\
 & Trend form & Additive linear ramp \\
\midrule

\multicolumn{3}{l}{\textbf{Shift Strength}} \\
\midrule
 & Strength levels & $\{1.0,\,0.7,\,0.5,\,0.35,\,0.25,\,0.18,\,0.12,\,0.08\}$ \\
 & Interpretation & $s=1$: strongest shift, $s\to0$: indistinguishable \\
\bottomrule
\end{tabular}
}
\end{table}

\input{figures/ar1_visualize}

%% file: figures/phi_raw_samples.tex

\begin{figure}[t]
\centering

\begin{subfigure}{0.48\linewidth}
    \centering
    \includegraphics[width=\linewidth]{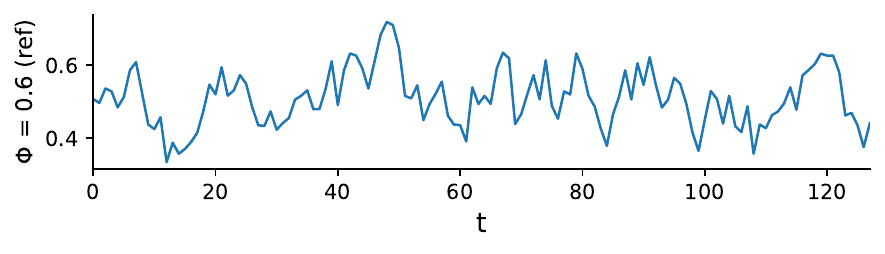}
    \caption{$\phi = 0.6$ (stationary)}
\end{subfigure}
\hfill
\begin{subfigure}{0.48\linewidth}
    \centering
    \includegraphics[width=\linewidth]{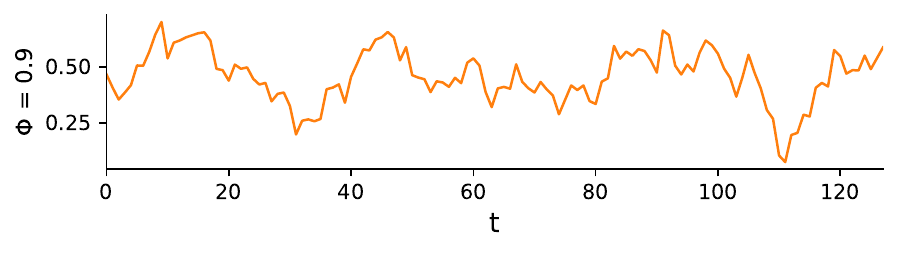}
    \caption{$\phi = 0.9$ (strong persistence)}
\end{subfigure}

\vspace{0.5em}

\begin{subfigure}{0.48\linewidth}
    \centering
    \includegraphics[width=\linewidth]{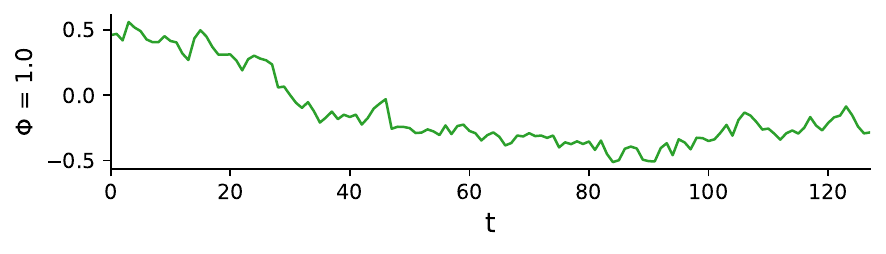}
    \caption{$\phi = 1.0$ (unit-root)}
\end{subfigure}
\hfill
\begin{subfigure}{0.48\linewidth}
    \centering
    \includegraphics[width=\linewidth]{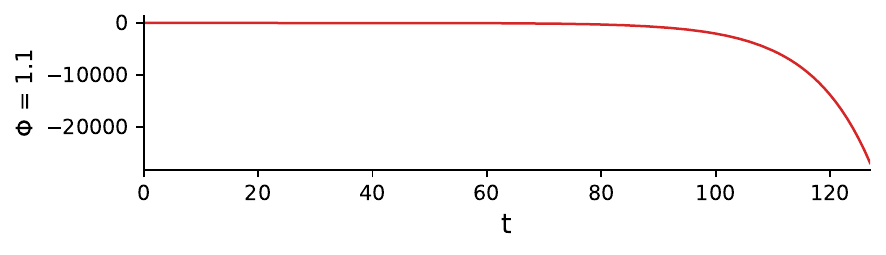}
    \caption{$\phi = 1.1$ (explosive)}
\end{subfigure}

\caption{
Representative AR(1) windows with increasing persistence.
As $\phi$ increases from $0.6$ to $1.1$, the process transitions from a mean-reverting regime to a unit-root boundary and eventually to an explosive regime.
Unlike distributional shifts, these changes modify temporal dependence rather than introducing explicit within-window shifts.
}
\label{fig:phi_raw_examples}
\end{figure}

%% file: figures/ar1_visualize.tex
\begin{figure}[t]
\centering
\includegraphics[width=0.95\linewidth]{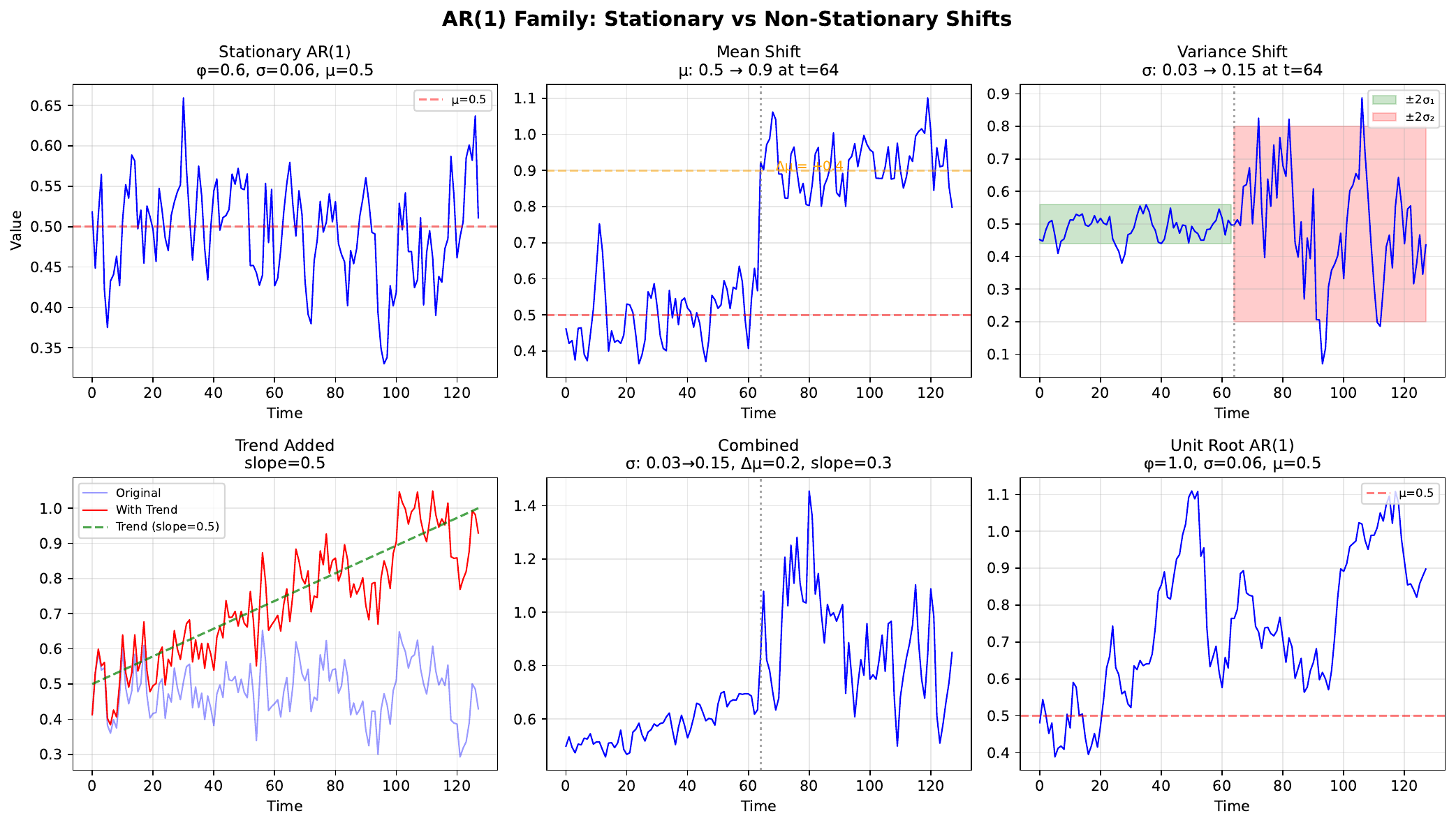}
\caption{
Representative AR(1) windows used for distributional non-stationarity experiments ($L=128$):
stationary, mean shift, variance shift, and trend.
}
\label{fig:ar1_synth_vis}
\end{figure}

%% file: appendix/2_model_explanation.tex
\section{Models}
\label{sec:appendix_models}

We evaluate three representative time series foundation models (TSFMs): Chronos2 \citep{ansari2025chronos2}, MOMENT \citep{goswami2024moment}, and TOTEM \citep{talukder2024totem}.
These models were selected to span diverse architectural paradigms and training objectives while enabling consistent extraction of window-level embeddings. Additionally, we evaluate two non-TSFM baselines \textit{Stats-LR} and \textit{Stats+Dynamics-LR}. 

\subsection{Chronos2}
Chronos2 \citep{ansari2025chronos2} is a pretrained time series foundation model designed for universal forecasting across univariate and multivariate settings.\footnote{\url{https://github.com/amazon-science/chronos-forecasting}}
It extends the Chronos family with group-attention mechanisms that enable cross-series information sharing and in-context learning.
The model processes normalized input sequences by segmenting them into patches, mapping them to embeddings, and applying a transformer stack with alternating time and group attention layers.

This design supports strong zero-shot forecasting and cross-domain generalization by leveraging shared temporal patterns across related series.
Because Chronos2 provides fixed-length representations for input windows, it is well suited for embedding-based analysis.

\paragraph{Usage in this work.}
We use Chronos2 as an encoder to obtain window embeddings via mean pooling over token representations.

\subsection{MOMENT}
MOMENT \citep{goswami2024moment} is a family of open-source foundation models for general-purpose time series analysis, trained via large-scale multi-dataset pretraining.\footnote{\url{https://github.com/moment-timeseries-foundation-model/moment}}
It learns representations through self-supervised objectives such as masked reconstruction, enabling a single model to support forecasting, classification, anomaly detection, and imputation tasks.

By reconstructing masked segments, MOMENT learns embeddings that preserve salient temporal structure while promoting invariances to nuisance variations.
This reconstruction-oriented objective makes MOMENT particularly suitable for studying which aspects of non-stationarity are preserved or suppressed in representation space.

\paragraph{Usage in this work.}
We use MOMENT as an embedder only, removing the reconstruction head and extracting latent representations directly from the encoder.

\subsection{TOTEM}
TOTEM (TOkenized Time Series EMbeddings) \citep{talukder2024totem} is a generalist time series foundation model based on discrete tokenization of time series data.\footnote{\url{https://github.com/SaberaTalukder/TOTEM}}
It employs a VQ-VAE-style architecture with convolutional encoders and a learned codebook to produce discrete token representations that can be used across forecasting, anomaly detection, and imputation tasks.

By representing time series as discrete tokens, TOTEM enables cross-domain training and strong zero-shot performance across diverse tasks. 
Its forecasting-oriented training encourages representations that emphasize temporal dynamics and structural patterns.

\paragraph{Usage in this work.}
We use the forecasting variant of TOTEM and extract embeddings from its encoder to obtain window-level representations.

\subsection{Baselines}


Our two baselines derive features from the time series without using TSFMs. The first baseline, \textit{Stats-LR}, includes the following features: quantiles, mean, standard deviation, minimum, maximum, range, inter-quartile range, absolute mean, root mean square, mean square root, skewness, and kurtosis. The second baseline, \textit{Stats+Dynamics-LR}, includes those same features in addition to: mean of first differences, standard deviation of first differences, slope of linear regression, and autocorrelations at lags 1, 2, and 3. These features train a logistic regression model to perform shift type classification. 

\subsection{Why These TSFMs?}
We selected these models for three primary reasons.

\paragraph{(1) Comparable embedding extraction.}
All three models provide encoder outputs that can be converted into fixed-length window embeddings without task-specific fine-tuning, enabling consistent representation-level diagnostics.

\paragraph{(2) Diverse inductive biases.}
Chronos2 emphasizes cross-series forecasting and probabilistic modeling, MOMENT focuses on reconstruction-driven representation learning, and TOTEM leverages discrete tokenization and forecasting objectives.
This diversity allows us to examine how training objectives shape the encoding of non-stationarity.

\paragraph{(3) Relevance to modern TSFMs.}
These models are representative of current design trends in time series foundation models and are widely used across forecasting and representation-learning tasks.

%% file: appendix/4_results.tex
\section{Experiment Results}
\label{app:results}
\subsection{Temporal Persistence Regression}
\label{app:phi_regression}

To complement the distributional shift analysis, we evaluate how well TSFM embeddings preserve temporal persistence.
Specifically, we regress the AR(1) coefficient $\phi$ from window-level embeddings and measure prediction accuracy.

For each window, we generate AR(1) sequences with $\phi \in [0.3, 1.1]$ and train a linear regressor to predict $\phi$ from the embedding.
We report mean absolute error (MAE), Pearson correlation ($r$), and coefficient of determination ($R^2$).

\subsubsection{Results}

\begin{table}[t]
\centering
\caption{Regression performance for predicting AR(1) persistence $\phi$ from embeddings. Lower MAE and higher $r$ and $R^2$ indicate better preservation of temporal dependence.}
\label{tab:phi_regression}
\setlength{\tabcolsep}{10pt}
\renewcommand{\arraystretch}{1.2}
\begin{tabular}{lccc}
\toprule
\textbf{Model} & \textbf{MAE} $\downarrow$ & \textbf{Pearson $r$} $\uparrow$ & \textbf{$R^2$} $\uparrow$ \\
\midrule
Chronos2 & 0.041 & 0.982 & 0.964 \\
MOMENT   & 0.058 & 0.965 & 0.931 \\
TOTEM    & 0.121 & 0.812 & 0.659 \\
Stats-LR & 0.089 & 0.901 & 0.811 \\
Stats+Dynamics-LR & 0.061 & 0.954 & 0.910 \\
\bottomrule
\end{tabular}
\end{table}

\paragraph{Interpretation.}
Chronos2 embeddings preserve temporal persistence most faithfully, achieving the lowest MAE and highest correlation.
MOMENT also captures persistence well, though with slightly reduced accuracy.
In contrast, TOTEM exhibits substantially weaker alignment with $\phi$, indicating that its embeddings encode persistence less explicitly.

Interestingly, the statistics-based baselines achieve competitive performance, suggesting that persistence can be partially recovered from low-order dynamics.
However, TSFMs—particularly Chronos2—provide a more precise and stable encoding of temporal dependence.
\subsection{Sequence Length Ablation}
\label{app:seq_len_ablation}

To test whether shift-type decodability depends on the choice of window length, we repeat the shift-type linear probing experiment across multiple sequence lengths $L \in \{64,128,256,512\}$.
We report Macro-F1 for representative shift strengths $s \in \{1.0, 0.25, 0.12\}$.
The default setting used in the main paper is $L=128$.

\subsubsection{Fixed Persistence ($\phi=0.6$)}

Table~\ref{tab:seq_len_phi_fixed} summarizes Macro-F1 across sequence lengths under fixed persistence ($\phi=0.6$).
Longer windows consistently improve separability for all methods, reflecting the benefit of additional temporal context.
Importantly, the qualitative model ranking is unchanged across $L$, and the weak-shift regime ($s=0.12$) continues to reveal substantial robustness differences across representations.

\begin{table}[t]
\centering
\caption{Sequence-length ablation for shift-type probing under fixed persistence ($\phi=0.6$). Values are Macro-F1 (mean$\pm$std over 5 seeds).}
\label{tab:seq_len_phi_fixed}
\setlength{\tabcolsep}{7pt}
\renewcommand{\arraystretch}{1.15}
\begin{tabular}{l|l|cccc}
\toprule
\textbf{Strength} & \textbf{Model} & $\mathbf{L=64}$ & $\mathbf{L=128}$ & $\mathbf{L=256}$ & $\mathbf{L=512}$ \\
\midrule
\multicolumn{6}{l}{\textit{\textbf{Strong Shifts} ($s=1.0$)}} \\
\midrule
\multirow{3}{*}{\textbf{TSFMs}} & Chronos2 & 0.891$\pm$0.007 & 0.924$\pm$0.005 & 0.948$\pm$0.005 & 0.969$\pm$0.002 \\
& MOMENT & 0.859$\pm$0.007 & 0.891$\pm$0.004 & 0.914$\pm$0.004 & 0.933$\pm$0.003 \\
& TOTEM & 0.711$\pm$0.005 & 0.741$\pm$0.005 & 0.760$\pm$0.007 & 0.780$\pm$0.008 \\
\midrule
\multirow{2}{*}{\textbf{Baselines}} & Stats-LR & 0.927$\pm$0.002 & 0.959$\pm$0.001 & 0.978$\pm$0.001 & 0.990$\pm$0.001 \\
& Stats+Dynamics-LR & 0.962$\pm$0.002 & 0.975$\pm$0.002 & 0.983$\pm$0.001 & 0.990$\pm$0.000 \\
\midrule
\multicolumn{6}{l}{\textit{\textbf{Moderate Shifts} ($s=0.25$)}} \\
\midrule
\multirow{3}{*}{\textbf{TSFMs}} & Chronos2 & 0.560$\pm$0.007 & 0.680$\pm$0.005 & 0.801$\pm$0.005 & 0.891$\pm$0.003 \\
& MOMENT & 0.523$\pm$0.006 & 0.620$\pm$0.006 & 0.713$\pm$0.006 & 0.789$\pm$0.009 \\
& TOTEM & 0.379$\pm$0.008 & 0.436$\pm$0.007 & 0.501$\pm$0.011 & 0.577$\pm$0.003 \\
\midrule
\multirow{2}{*}{\textbf{Baselines}} & Stats-LR & 0.435$\pm$0.007 & 0.501$\pm$0.006 & 0.569$\pm$0.007 & 0.635$\pm$0.005 \\
& Stats+Dynamics-LR & 0.530$\pm$0.005 & 0.617$\pm$0.003 & 0.692$\pm$0.002 & 0.757$\pm$0.005 \\
\midrule
\multicolumn{6}{l}{\textit{\textbf{Weak Shifts} ($s=0.12$)}} \\
\midrule
\multirow{3}{*}{\textbf{TSFMs}} & Chronos2 & 0.347$\pm$0.005 & 0.396$\pm$0.007 & 0.500$\pm$0.007 & 0.643$\pm$0.003 \\
& MOMENT & 0.339$\pm$0.008 & 0.373$\pm$0.011 & 0.442$\pm$0.003 & 0.501$\pm$0.005 \\
& TOTEM & 0.251$\pm$0.007 & 0.268$\pm$0.005 & 0.297$\pm$0.006 & 0.349$\pm$0.007 \\
\midrule
\multirow{2}{*}{\textbf{Baselines}} & Stats-LR & 0.293$\pm$0.007 & 0.323$\pm$0.012 & 0.356$\pm$0.013 & 0.377$\pm$0.006 \\
& Stats+Dynamics-LR & 0.345$\pm$0.005 & 0.384$\pm$0.005 & 0.442$\pm$0.006 & 0.503$\pm$0.008 \\
\bottomrule
\end{tabular}
\end{table}

\subsubsection{Random Persistence ($\phi \sim U(0.3,0.9)$)}

Table~\ref{tab:seq_len_phi_random} repeats the same evaluation under random persistence.
The same qualitative trends hold: longer windows improve decodability, while the relative robustness ordering across representations remains stable.
This supports the conclusion that the observed failure modes are not artifacts of a particular window length or a fixed choice of $\phi$.

\begin{table}[t]
\centering
\caption{Sequence-length ablation for shift-type probing under random persistence ($\phi \sim U(0.3,0.9)$). Values are Macro-F1 (mean$\pm$std over 5 seeds).}
\label{tab:seq_len_phi_random}
\setlength{\tabcolsep}{7pt}
\renewcommand{\arraystretch}{1.15}
\begin{tabular}{l|l|cccc}
\toprule
\textbf{Strength} & \textbf{Model} & $\mathbf{L=64}$ & $\mathbf{L=128}$ & $\mathbf{L=256}$ & $\mathbf{L=512}$ \\
\midrule
\multicolumn{6}{l}{\textit{\textbf{Strong Shifts} ($s=1.0$)}} \\
\midrule
\multirow{3}{*}{\textbf{TSFMs}} & Chronos2 & 0.866$\pm$0.005 & 0.898$\pm$0.003 & 0.927$\pm$0.003 & 0.953$\pm$0.003 \\
& MOMENT & 0.836$\pm$0.005 & 0.873$\pm$0.005 & 0.903$\pm$0.004 & 0.927$\pm$0.004 \\
& TOTEM & 0.702$\pm$0.006 & 0.727$\pm$0.013 & 0.752$\pm$0.006 & 0.767$\pm$0.008 \\
\midrule
\multirow{2}{*}{\textbf{Baselines}} & Stats-LR & 0.899$\pm$0.003 & 0.922$\pm$0.002 & 0.940$\pm$0.002 & 0.952$\pm$0.002 \\
& Stats+Dynamics-LR & 0.929$\pm$0.002 & 0.948$\pm$0.002 & 0.961$\pm$0.001 & 0.969$\pm$0.001 \\
\midrule
\multicolumn{6}{l}{\textit{\textbf{Moderate Shifts} ($s=0.25$)}} \\
\midrule
\multirow{3}{*}{\textbf{TSFMs}} & Chronos2 & 0.541$\pm$0.007 & 0.662$\pm$0.007 & 0.788$\pm$0.004 & 0.871$\pm$0.004 \\
& MOMENT & 0.508$\pm$0.006 & 0.599$\pm$0.006 & 0.687$\pm$0.004 & 0.760$\pm$0.006 \\
& TOTEM & 0.346$\pm$0.006 & 0.378$\pm$0.006 & 0.430$\pm$0.005 & 0.484$\pm$0.007 \\
\midrule
\multirow{2}{*}{\textbf{Baselines}} & Stats-LR & 0.386$\pm$0.010 & 0.421$\pm$0.010 & 0.448$\pm$0.008 & 0.470$\pm$0.010 \\
& Stats+Dynamics-LR & 0.476$\pm$0.009 & 0.552$\pm$0.006 & 0.616$\pm$0.004 & 0.668$\pm$0.006 \\
\midrule
\multicolumn{6}{l}{\textit{\textbf{Weak Shifts} ($s=0.12$)}} \\
\midrule
\multirow{3}{*}{\textbf{TSFMs}} & Chronos2 & 0.345$\pm$0.006 & 0.398$\pm$0.007 & 0.493$\pm$0.007 & 0.630$\pm$0.003 \\
& MOMENT & 0.340$\pm$0.006 & 0.374$\pm$0.009 & 0.431$\pm$0.004 & 0.479$\pm$0.009 \\
& TOTEM & 0.244$\pm$0.004 & 0.249$\pm$0.007 & 0.268$\pm$0.005 & 0.285$\pm$0.004 \\
\midrule
\multirow{2}{*}{\textbf{Baselines}} & Stats-LR & 0.270$\pm$0.007 & 0.281$\pm$0.008 & 0.302$\pm$0.011 & 0.321$\pm$0.015 \\
& Stats+Dynamics-LR & 0.331$\pm$0.007 & 0.354$\pm$0.005 & 0.394$\pm$0.004 & 0.430$\pm$0.009 \\
\bottomrule
\end{tabular}
\end{table}

\input{table/tab_macro_f1_main}

\paragraph{Interpretation.}
Chronos2 embeddings preserve temporal persistence most faithfully, achieving the lowest MAE and highest correlation.
MOMENT also captures persistence well, though with slightly reduced accuracy.
In contrast, TOTEM exhibits substantially weaker alignment with $\phi$, indicating that its embeddings encode persistence less explicitly.

\subsection{Confusion Matrix Analysis}
\label{app:confusion}

To better understand model-specific failure modes, we examine confusion matrices for shift-type classification at representative shift strengths.
Rows correspond to true labels and columns to predicted labels.

Figures~\ref{fig:confusion_fixed} and~\ref{fig:confusion_random} show results for fixed persistence ($\phi=0.6$) and random persistence ($\phi \sim U(0.3,0.9)$), respectively.

\input{figures/confusion_fixed}
\input{figures/confusion_random}

%% file: table/tab_macro_f1_main.tex
\begin{table}[t]
\centering
\caption{Shift-type linear probing under varying shift strengths ($\phi=0.6$ and $\phi \sim U(0.3, 0.9)$).}
\label{tab:macro_f1_integrated}
\setlength{\tabcolsep}{8pt} 
\renewcommand{\arraystretch}{1.2}
\begin{tabular}{l|l|ccc} 
\toprule
\textbf{Type} & \textbf{Model} & $\mathbf{s=1.0}$ & $\mathbf{s=0.25}$ & $\mathbf{s=0.12}$ \\
\midrule
\multicolumn{5}{l}{\textit{\textbf{Fixed Shift} ($\phi=0.6$)}} \\
\midrule
\multirow{3}{*}{\textbf{TSFMs}} & Chronos2 & 0.924 $\pm$ 0.005 & \textbf{0.680 $\pm$ 0.005} & \textbf{0.396 $\pm$ 0.007} \\
& MOMENT & 0.891 $\pm$ 0.004 & 0.620 $\pm$ 0.006 & 0.373 $\pm$ 0.011 \\
& TOTEM & 0.741 $\pm$ 0.005 & 0.436 $\pm$ 0.007 & 0.268 $\pm$ 0.005 \\
\midrule
\multirow{2}{*}{\textbf{Baselines}} & Stats-LR & 0.959 $\pm$ 0.001 & 0.501 $\pm$ 0.006 & 0.323 $\pm$ 0.012 \\
& Stats+Dynamics-LR & \textbf{0.975 $\pm$ 0.002} & 0.617 $\pm$ 0.003 & 0.384 $\pm$ 0.005 \\
\midrule
\midrule
\multicolumn{5}{l}{\textit{\textbf{Random Shift} ($\phi \sim U(0.3,0.9)$)}} \\
\midrule
\multirow{3}{*}{\textbf{TSFMs}} & Chronos2 & 0.898 $\pm$ 0.003 & \textbf{0.662 $\pm$ 0.007} & \textbf{0.398 $\pm$ 0.007} \\
& MOMENT & 0.873 $\pm$ 0.005 & 0.599 $\pm$ 0.006 & 0.374 $\pm$ 0.009 \\
& TOTEM & 0.727 $\pm$ 0.013 & 0.378 $\pm$ 0.006 & 0.249 $\pm$ 0.007 \\
\midrule
\multirow{2}{*}{\textbf{Baselines}} & Stats-LR & 0.922 $\pm$ 0.002 & 0.421 $\pm$ 0.010 & 0.281 $\pm$ 0.008 \\
& Stats+Dynamics-LR & \textbf{0.948 $\pm$ 0.002} & 0.552 $\pm$ 0.006 & 0.354 $\pm$ 0.005 \\
\bottomrule
\end{tabular}
\end{table}


%% file: figures/confusion_fixed.tex
\begin{figure*}[t]
\centering
\includegraphics[width=0.95\linewidth]{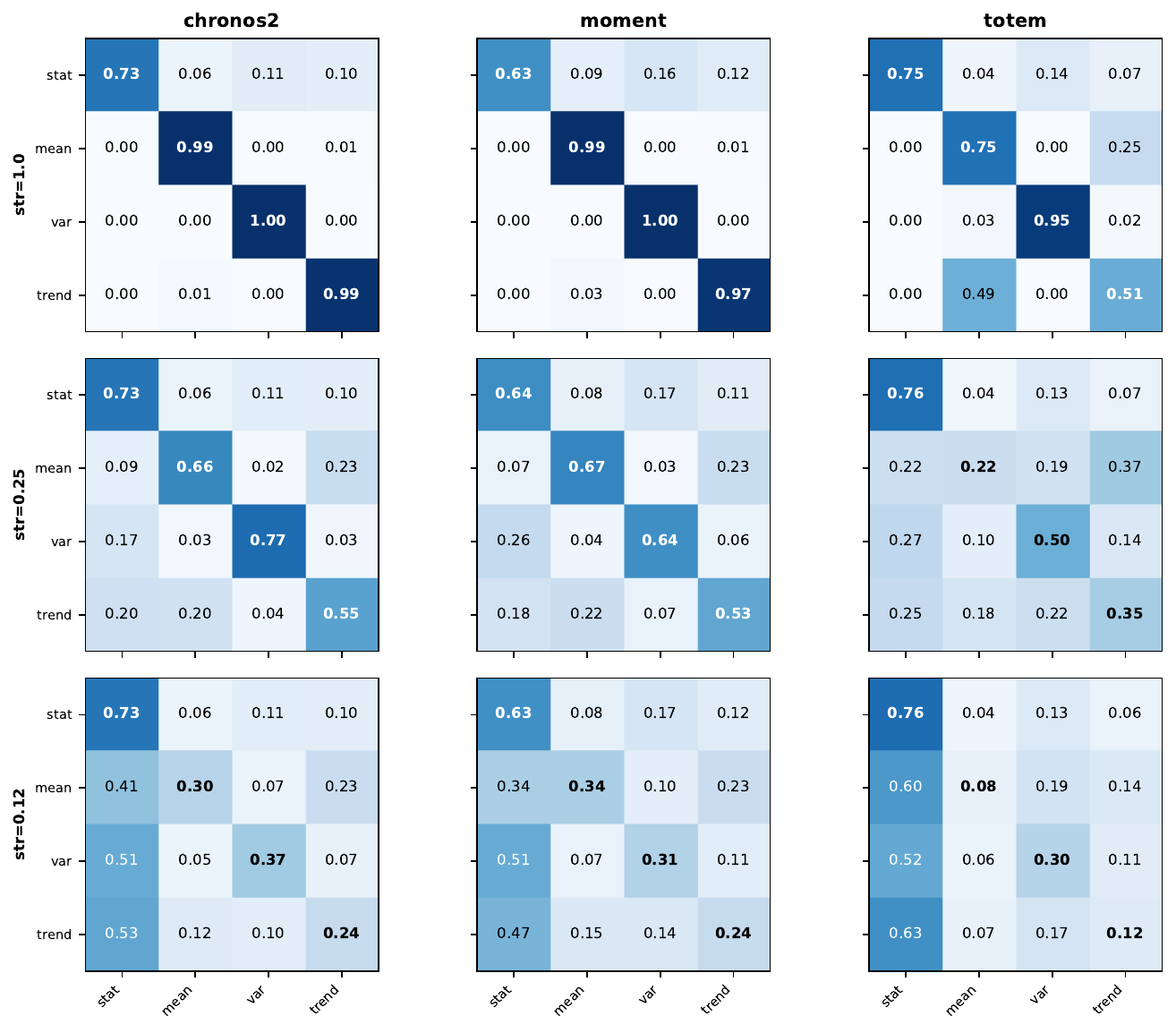}
\caption{
Confusion matrices for shift-type probing at sequence length $L=128$ with fixed persistence ($\phi=0.6$).
As shift strength decreases (top to bottom), errors increase and reveal model-specific failure modes.
Chronos2 and MOMENT retain strong diagonal structure for variance shifts, whereas TOTEM exhibits substantial collapse of mean shifts into the stationary class.
}
\label{fig:confusion_fixed}
\end{figure*}

%% file: figures/confusion_random.tex
\begin{figure*}[t]
\centering
\includegraphics[width=0.95\linewidth]{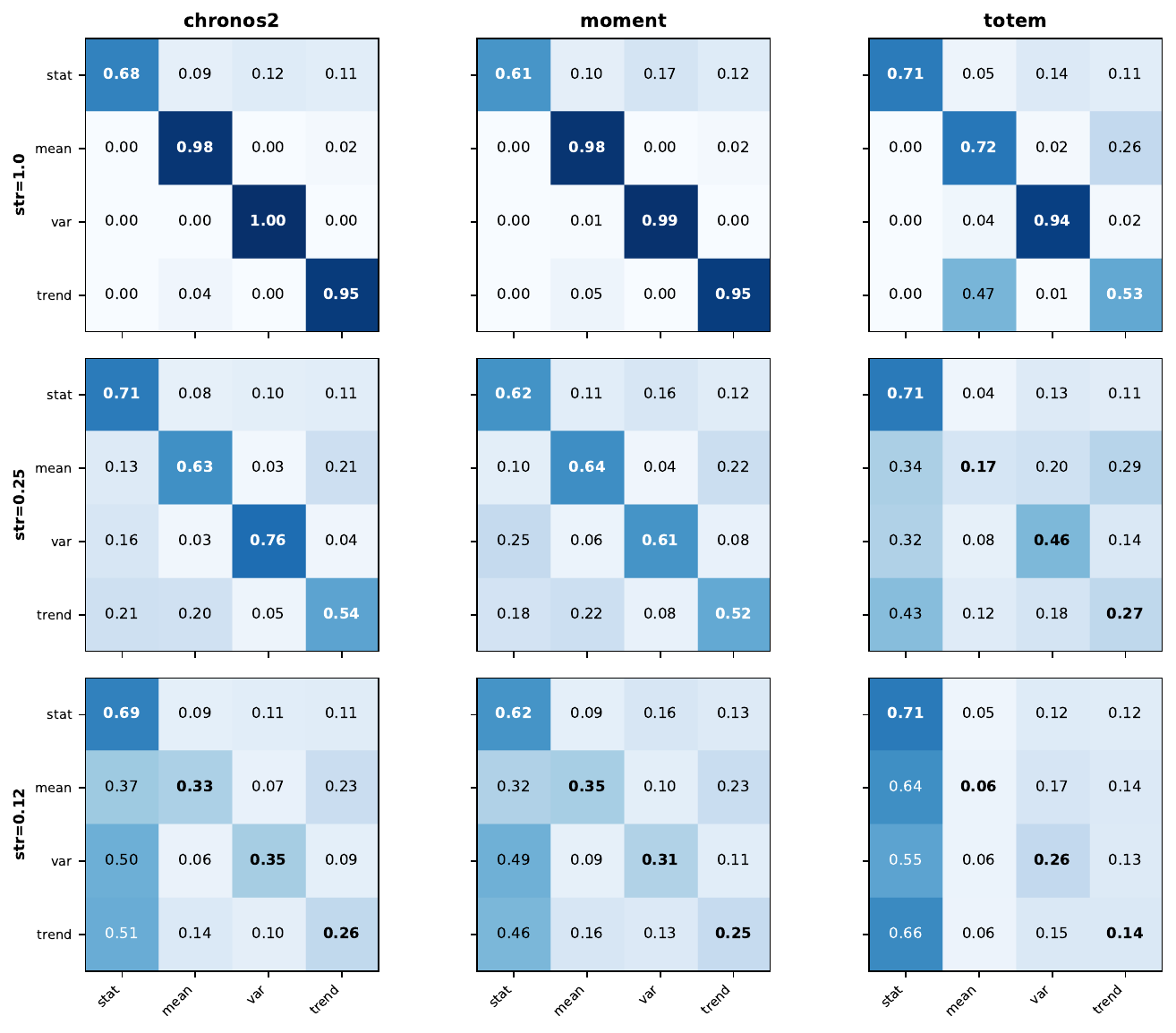}
\caption{
Confusion matrices under random persistence ($\phi \sim U(0.3,0.9)$).
Qualitative failure patterns remain consistent with the fixed-$\phi$ setting, indicating that shift-type confusion is driven by distributional structure rather than persistence.
}
\label{fig:confusion_random}
\end{figure*}

%% file: appendix/3_background.tex
\section{Related Work: Perspectives on Non-Stationarity}
\label{sec:related_work}

Non-stationarity has long been recognized as a central challenge in time series analysis, yet its interpretation and treatment vary substantially across the literature.
In classical statistics, non-stationarity is formalized through precise definitions, such as violations of weak stationarity or the presence of unit roots~\citep{ADF, kpss, pptest, sjosten2022comparative}.
In contrast, many modern deep learning approaches adopt a broader and often less explicit view, frequently using non-stationarity as an umbrella term for diverse forms of temporal variation.
This section reviews how non-stationarity has been conceptualized and addressed in recent time series modeling work, with a focus on the assumptions implicit in these approaches.

\paragraph{Non-stationarity as distribution shift.}
A large body of recent work implicitly equates non-stationarity with distributional change over time.
For example, several studies characterize non-stationary time series as those exhibiting continuously changing statistical properties or joint distributions, which are argued to hinder predictability.
Within this perspective, non-stationarity is treated primarily as a nuisance factor that should be mitigated to simplify learning.
Normalization-based techniques are commonly proposed to stabilize time series by removing shifts in mean, variance, or scale~\citep{dain, kim2021reversible}.
More structured approaches explicitly model time-varying normalization and denormalization processes to compensate for distributional drift during forecasting \citep{liu2022non}.

Related ideas have also been explored in representation-space analyses.
For instance, Dish-TS \citep{dishts} distinguishes between \emph{intra-space shift}, referring to temporal changes within a single representation space, and \emph{inter-space shift}, describing misalignment across representations learned under different temporal regimes.
These formulations further reinforce the view of non-stationarity as a form of distribution shift that disrupts stable representation learning.

\paragraph{Questioning over-stabilization.}
While normalization-based approaches have shown empirical success, a growing line of work challenges the assumption that non-stationarity should always be removed.
Several recent studies argue that aggressive stabilization may inadvertently discard informative temporal structure \citep{liu2022non, wang2024timemixer++, liu2025timebridge}.
From this perspective, non-stationarity is not merely a source of noise but may encode meaningful dynamics that are essential for predictive tasks.

\paragraph{Positioning of this work.}
Our work differs from the above lines of research in both scope and intent.
Rather than advocating for or against normalization, or proposing a new architectural solution, we adopt a diagnostic perspective.
We ask how different meanings of non-stationarity—distributional shifts versus temporal dependence violations—manifest in the embedding spaces of time series foundation models.
By disentangling these notions and analyzing their representation-level effects under multiple metrics, we aim to clarify what is preserved, suppressed, or entangled in learned embeddings, and how prior assumptions about non-stationarity shape model behavior.